\documentclass{article}
\usepackage{iclr2027_conference,times}
\iclrfinalcopy
\usepackage[utf8]{inputenc}
\usepackage[T1]{fontenc}
\usepackage{amsmath,amssymb,booktabs,graphicx}
\usepackage{float}
\usepackage{placeins}
\usepackage{needspace}
\usepackage{longtable}
\usepackage{microtype}
\usepackage{xcolor}
\usepackage{hyperref}
\usepackage{url}
\hypersetup{pdfauthor={Tong Zhang, Yuxuan Cheng, Yun Peng},pdfsubject={Preprint},pdftitle={InfiniteWorld: Separating Visual Evidence from Familiar Physics},colorlinks=true,linkcolor=black,citecolor=black,urlcolor=black}
\newcommand{\benchmark}{\textsc{InfiniteWorld}}

\title{InfiniteWorld: Separating Visual Evidence\\from Familiar Physics}
\author{Tong Zhang\textsuperscript{1} \quad Yuxuan Cheng\textsuperscript{2} \quad Yun Peng\textsuperscript{1}\\
{\normalfont\textsuperscript{1}Fudan University}\\
{\normalfont\textsuperscript{2}The Chinese University of Hong Kong, Shenzhen}\\
{\normalfont\small\href{mailto:tongzhang25@m.fudan.edu.cn}{\texttt{tongzhang25@m.fudan.edu.cn}} \quad
\href{mailto:hanxuanwxss@gmail.com}{\texttt{hanxuanwxss@gmail.com}} \quad
\href{mailto:yunp@fudan.edu.cn}{\texttt{yunp@fudan.edu.cn}}}}
\begin{document}
\maketitle
\pagestyle{plain}
\begin{abstract}
Physical reasoning requires adapting familiar expectations to the dynamics of the observed world. Existing benchmarks establish broad physical capabilities, yet an overall score leaves unclear whether answers follow visual evidence when familiar rules no longer apply. We introduce \benchmark{}, a benchmark that pairs physical-reasoning questions with simulator videos whose physical parameters are systematically altered. Its generation pipeline connects simulation, scene screening, paired video translation, question construction, and input checks. Five tasks assess law identification, parameter estimation, temporal reasoning, analytic counterfactual comparison, and law induction. The evaluation defines five matched input settings for each question, comprising text-only evaluation and ordered or shuffled observations from simulator and translated videos. We evaluate a diverse set of video-language models under these matched settings. The results expose a persistent preference for familiar physical rules. Frames improve performance for most models, but gains differ across tasks. Strong accuracy on Earth-normal questions coexists with poor recognition of altered dynamics, and Earth-normal answers remain common when the observed physics changes. Task scores and semantic agreement further distinguish matching physical factors from matching their amount, and correct answers from repeated errors. \benchmark{} makes these behaviors measurable through paired changes in the evidence supplied for the same question, separating visual evidence from familiar physics.
\end{abstract}
\begin{figure}[H]
\centering
\includegraphics[width=\linewidth]{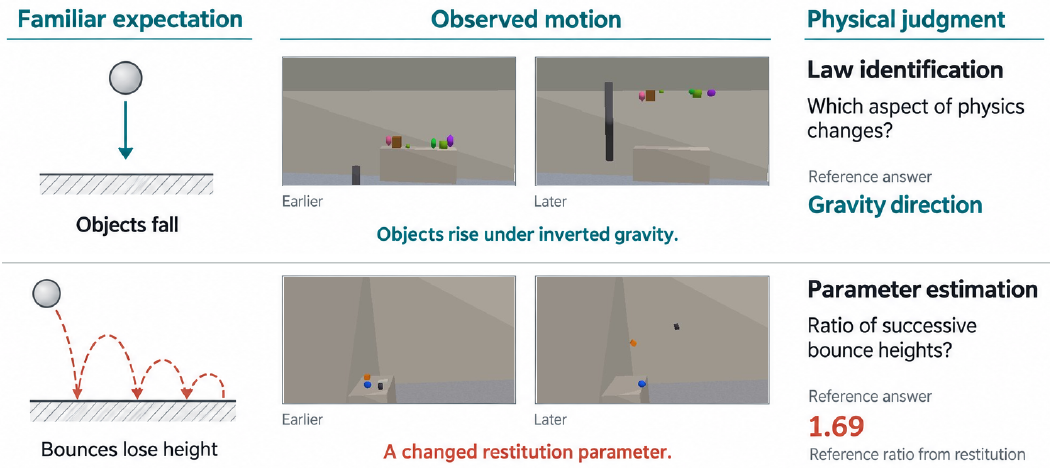}
\caption{\textbf{When familiar expectations conflict with observed motion.} Earth-normal expectations (left, schematic) contrast with recorded altered dynamics (center) and two physical judgments (right). Frames are fixed crops from the evaluated simulator videos in chronological order. The numeric reference is $e^2$, where $e$ is the ratio of outgoing to incoming normal collision speed. Questions and answers are shortened for display. Appendix O includes the full questions and paired renderings. }
\label{fig:teaser}
\label{fig:task_examples}
\end{figure}

\raggedbottom
\section{Introduction}
\label{sec:intro}
\begin{quote}\small\itshape
``Look upon the things of this world, and you will see convention; invert them, and you will see the truth.''
\end{quote}

Physical reasoning is an important capability for agents to operate correctly in the
real world. To perceive and predict physical events, a vision-language model
draws on two abilities. One is the familiar physical knowledge
learned during pretraining. The other is the interpretation of visual evidence in
the current question~\citep{battaglia2013,lake2017}. When the observed motion
matches the learned knowledge, the two abilities support the same answer.
When the physics is altered, they diverge, and the visual evidence and the
learned knowledge favor different answers. In practice, a model can often
produce a correct answer by reasoning from the question text and its
familiar physical knowledge alone, leaving the visual evidence unused. This shortcut often succeeds, because familiar physical rules describe
most situations. When an object is lighter or heavier than it looks, a
surface is more slippery than it appears, or an unfamiliar force drives the
motion, an answer from familiar physical knowledge alone describes what
usually happens rather than what is happening in the video. An agent that
acts on that answer fails in precisely these situations.
An evaluation should therefore measure both answer quality and the benefit of visual evidence.

Existing physical-reasoning benchmarks for vision-language models often summarize performance
through task accuracy. PhysBench evaluates object
properties, relationships, scene understanding, and
dynamics~\citep{physbench}. CLEVRER studies causal reasoning about object
interactions~\citep{clevrer}. Physion and Physion++ evaluate prediction
from motion and inference of hidden physical
properties~\citep{physion,physionpp}. On Earth-normal scenes, the same correct answer can follow visual evidence or familiar knowledge, so accuracy alone does not show which source supports the answer.
Others tests physics that
differs from familiar experience. IntPhys and PLATO test sensitivity to
violations of physical expectations~\citep{intphys,plato}. InPhyRe studies
learning of altered collision rules from
demonstrations~\citep{inphyre}. NewtonBench evaluates discovery of modified
equations through interaction~\citep{newtonbench}.

The benchmarks above either evaluate familiar physical scenes, where prior knowledge can yield the correct answer without attending to the video, or evaluate altered physics while also changing the question, reference answer, or interaction format. Such evaluations make it difficult to separate the contribution of visual evidence from that of familiar physical knowledge, and consequently to systematically assess models' ability to perceive and predict physical events from visual evidence.

To study this problem, we introduce \benchmark{}, a benchmark that pairs questions with simulated videos of Earth-normal and altered dynamics. It links each question to a simulated video and recorded physical state, including six physical factors:
gravity direction and strength, friction, restitution, air dissipation,
and chiral deflection. The last two introduce velocity-dependent damping
and motion-dependent sideways acceleration, respectively. The parameter values
vary, either alone or in seven selected two-factor
combinations, such as gravity direction with restitution. Scenes using
the Earth-normal reference settings are Earth-normal, while scenes with one or
two altered factors are altered. Together, these Earth-normal and altered scenes support five tasks that probe whether models can reason from observed motion, especially when it conflicts with familiar physical expectations.

To examine how visual evidence affects performance on these tasks, we evaluate each question under five matched input settings. These are Text only, Simulator ordered, Simulator shuffled, Translated ordered, and Translated shuffled. The question wording, answer options, and reference answer remain fixed across settings. Score differences give the visual gain, order effect, and rendering gap.

Across the evaluated models, Simulator ordered usually improves over
Text only, but the benefit differs by task. Law identification improves
consistently, while analytic counterfactual comparison shows both gains
and losses. For Qwen3.8-27B, Simulator ordered improves accuracy on
altered scenes over Text only, but Earth-normal answers still dominate.
The model remains much more accurate on Earth-normal questions than on
altered questions.

Our contributions are as follows.

\begin{itemize}
\setlength{\itemsep}{2pt}
\setlength{\parsep}{0pt}
\item \textbf{A benchmark of Earth-normal and altered dynamics.}
\benchmark{} links simulator videos to questions across five physical-reasoning tasks. Its construction links
simulated motion, paired simulator and translated videos, and
task-specific reference answers.

\item \textbf{A matched-input evaluation of visual evidence.}
Each question retains its wording, answer options, and reference answer
across five input settings. Paired comparisons measure changes in score with frame availability, frame order, and rendering.

\item \textbf{Findings on visual gains and Earth-normal errors.}
Simulator ordered usually improves the score over Text only, but gains
vary across tasks and Earth-normal answers remain common on altered
law-identification questions.
\end{itemize}

\section{Benchmark}
\label{sec:dataset}

\subsection{Scenes}
A scene is a physical configuration specified by object geometry, initial positions and velocities, and dynamics parameters. Earth-normal scenes use the Earth-normal reference settings. Altered scenes change at least one setting. The generator varies six factors. Gravity direction and strength specify gravitational acceleration. Friction sets resistance to sliding. Restitution is the ratio of outgoing to incoming normal speeds at a collision. Air dissipation introduces velocity-dependent damping. Chiral deflection adds a sideways acceleration whose orientation depends on motion direction.

A generator category specifies a factor or factor combination and its amount category. An amount category is a discrete level of the physical change. The collection covers Earth-normal settings, individual altered factors, and selected factor combinations. Scene templates and seeds vary object arrangements and initial conditions within these categories. A template can also affect the effective coefficients, so law induction uses the generator category as its target.

\subsection{Data Generation}
The pipeline begins with a factor or factor combination, parameter values, and a scene seed. Simulation produces the video and source trajectory used to construct reference answers. Scene and trajectory checks assess bounds, rendering contrast, object placement, and physical fits where defined. Accepted scenes supply simulator videos with a shared duration and frame rate; failed candidates are excluded. The generation tables report the retained scene categories.

A simulator video is the rendered video of a scene. A translated video is its paired output from depth-conditioned Cosmos translation~\citep{cosmostransfer,cosmoswfm}. Depth encodes the distance from the camera to scene surfaces. The initial rendering sizes are $832\times480$ and $768\times432$, respectively; evaluation uses a common frame size and sampling schedule. Shared answers support comparison across renderings, including any question-relevant content changes introduced by translation.

Questions and reference answers are constructed from source state and metadata. The scoring key stores reference options, numeric targets, and tolerances separately from the public questions. Identifier checks link every question to an accepted scene. Question regeneration checks the public questions and reference answers, while frame checks verify the common sampling schedule across both renderings. This connects scene acceptance, target construction, and visual coverage.

\subsection{Tasks}
Table~\ref{tab:tasks} summarizes the five tasks and their question counts. Multiple-choice questions request an option number, while parameter-estimation questions request a numeric answer. Each question has a reference answer derived from its scene. In Figure~\ref{fig:task_examples}, upward motion identifies the change in gravity direction; the parameter-estimation question asks for the ratio of successive bounce heights. The distinction is between identifying the altered factor and estimating its effect.

\begin{table}[t]
\centering\small
\caption{\textbf{Tasks.} Earth-normal counts questions whose designated simulator videos use Earth-normal parameters. Law induction designates altered reference videos, although its query can be Earth-normal. Temporal questions use observations spanning the entire video.}
\label{tab:tasks}
\begin{tabular}{@{}lrrp{5.15cm}@{}}
\toprule
Task & Questions & Earth-normal & Reference answer\\
\midrule
Law identification & 6,764 & 2,396 & One listed altered factor or no alteration.\\
Parameter estimation & 1,230 & 599 & A physical quantity or bounce-height ratio.\\
Temporal reasoning & 3,024 & 1,095 & Relative height at two times within the video.\\
Analytic counterfactual comparison & 2,715 & 1,198 & Height or distance under an analytic reference.\\
Law induction & 1,496 & 0 & Whether a query and two references share a generator category.\\
\midrule
Total & 15,229 & 5,288 & 3,382 distinct simulator videos.\\
\bottomrule
\end{tabular}
\end{table}

\paragraph{Law identification.}
The model identifies a physical factor that differs from the reference, or selects no alteration. Each video yields two four-option questions with the same answer in different arrangements. Every set includes the no-alteration answer. For a scene with two altered factors, one becomes the queried factor and the other is excluded from the distractors, giving one correct answer.

\paragraph{Parameter estimation.}
The model returns a gravity-related quantity or the ratio of consecutive bounce heights. Full-credit and half-credit tolerances measure numerical agreement with the reference value. The bounce-height ratio is one rebound's height divided by the preceding height and serves as an observable proxy for restitution.

\paragraph{Temporal reasoning.}
The model compares a tracked object's height across a fixed time interval. The reference answer follows displacement in the recorded trajectory. Supplied frames span the entire video, so Temporal reasoning compares positions within the observed sequence. Textual time queries must be aligned with frames presented without timestamp annotations.

\paragraph{Analytic counterfactual comparison.}
The model compares height or traveled distance with an Earth-normal reference under the same initial scene. Credit measures agreement with this analytic reference.

\paragraph{Law induction.}
The reference videos share an altered generator category. The model decides whether a query video belongs to that category. Match questions require Yes; nonmatching questions change the factor or factor combination, use Earth-normal physics, or change the amount category and require No. The task balances matching and nonmatching queries to test category matching.

\subsection{Answer Patterns}
Question wording and answer choices can favor some answers before any frames are supplied. Permuting options changes answer positions while retaining their semantic content. Text-only evaluation and task-specific partitions characterize these preferences. A template-based rule predicts an answer from repeated question wording; rules selected after inspecting the fixed labels provide descriptive references for the observed answer distribution (Figure~\ref{fig:template}).

\section{Experimental Setup}
\label{sec:method}

\begin{figure}[t]
\centering
\includegraphics[width=\linewidth]{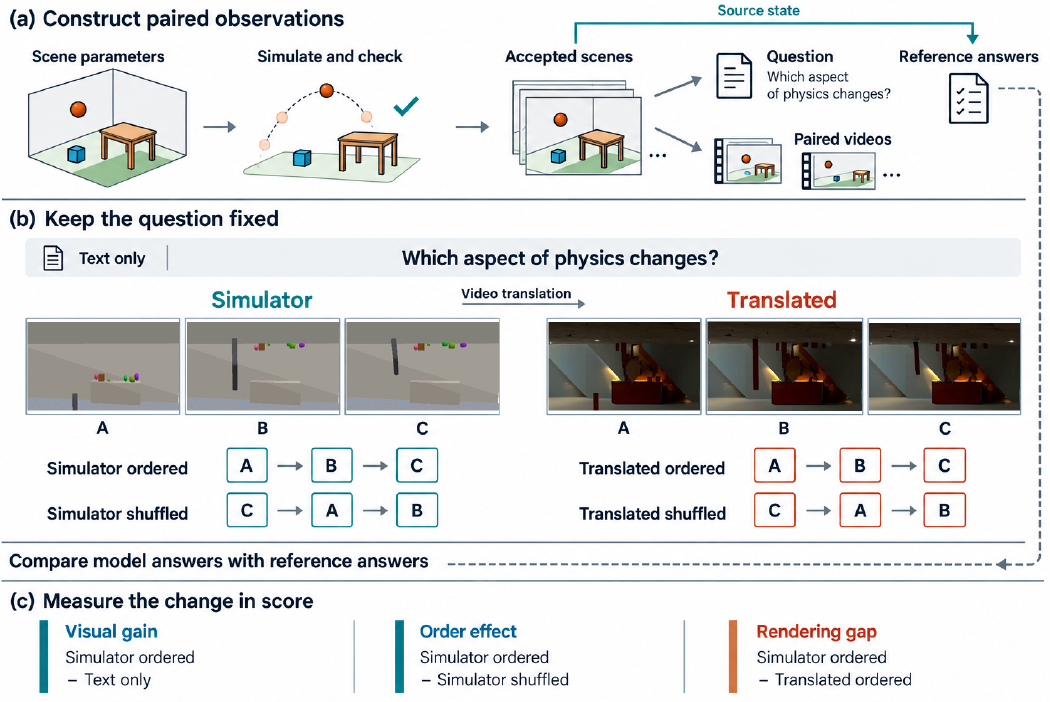}
\caption{\textbf{Construction and evaluation.} (a)~Simulation and scene checks produce accepted scenes, paired videos, questions, and reference answers. The construction drawings are schematic. (b)~The question and options stay identical across all input settings. Text only omits frames; the visual settings present simulator or translated frames in ordered or shuffled sequences. Shown observations are fixed crops from the law-identification example in Figure~\ref{fig:teaser}; letters identify their original positions, and the shuffled settings reorder the same frames. (c)~Paired scores measure visual gain, order effect, and rendering gap. Reference answers enter only scoring.}
\label{fig:framework}
\vspace{-15pt}
\end{figure}

\subsection{Input Settings}
Each question has five input settings that retain its wording, options, reference answer, and answer extractor. Text only supplies the question text and options. Simulator ordered supplies frames chronologically, while Simulator shuffled applies a deterministic permutation shared across models. Translated ordered and Translated shuffled use frames from the translated video in the corresponding orders. Law induction preserves the roles of its three videos and shuffles within each video.

Each video contributes uniformly sampled frames. Law induction supplies sequences for the reference and query videos. Frames retain their aspect ratio within a padded canvas. Prompts contain video markers, frames when applicable, and the public question, without timestamps or construction metadata. Reference answers are reserved for scoring.

\subsection{Metrics}
The score measures answer quality across the five tasks. The answer extractor returns an option number or numeric estimate. Multiple-choice credit $s_i$ is one for the reference option and zero otherwise. Numeric credit also rewards estimates within a wider tolerance, separating approximate estimates from larger errors. For estimate $\hat y_i$, reference value $y_i$, and tolerances $\tau_i^{(1)}$ and $\tau_i^{(1/2)}$,
\begin{equation}
 s_i=\begin{cases}
 1,&|\hat y_i-y_i|\leq\tau_i^{(1)},\\
 \tfrac12,&\tau_i^{(1)}<|\hat y_i-y_i|\leq\tau_i^{(1/2)},\\
 0,&\text{otherwise}.
 \end{cases}
\label{eq:score}
\end{equation}
Unextractable answers receive zero credit. For setting $a$, let $S_a=N^{-1}\sum_i s_i^a$ be the mean credit, where $N$ is the number of questions in the full set. The reported score is $100S_a$ (\%). Task scores use their task denominators, and the macro score averages the five task scores equally. Table~\ref{tab:main} reports the full-set score, weighting each question equally.

Correctness alone leaves the contribution of observations unresolved. Visual gain compares the simulator-ordered and text-only scores. The order effect compares the same frames chronologically and shuffled, and the rendering gap compares Simulator ordered and Translated ordered. For complete evaluations of the same questions,
\begin{align}
\Delta_{\mathrm{vis}} &=100(S_{\mathrm{simulator,ordered}}-S_{\mathrm{text-only}}),\\
\Delta_{\mathrm{order}} &=100(S_{\mathrm{simulator,ordered}}-S_{\mathrm{simulator,shuffled}}),\\
\Delta_{\mathrm{render}} &=100(S_{\mathrm{simulator,ordered}}-S_{\mathrm{translated,ordered}}).
\label{eq:contrasts}
\end{align}
The score differences are computed before rounding and expressed in percentage points. Positive values respectively indicate benefit from frames, benefit from chronological order, and a higher simulator-ordered score.

The remaining errors require task-specific measurements. The Earth-normal error rate is the fraction of altered law-identification questions answered with the Earth-normal answer. It quantifies the particular error that overall accuracy cannot distinguish from other wrong choices. Semantic agreement is the fraction of simulator videos whose two option arrangements yield the same valid answer text. Partitioning agreement into correct and wrong pairs separates stable correct judgments from repeated mistakes. Law-induction accuracy is reported for four question groups. In Match, the query belongs to the same generator category as both reference videos, so the correct answer is Yes. The No groups are Different factors, Earth-normal, and Different amount. This split distinguishes matching the factors from matching their amount and exposes Yes/No response preferences.

\subsection{Models}
Table~\ref{tab:main} groups the evaluated models by developer and output token limit. Each reported cell contains every assigned question. Decoding is greedy for the smaller output token limit, while the larger limit requests zero temperature. Scores and maxima are grouped by output token limit, which remains fixed across a model's input settings. Models receive no task-specific fine-tuning. Unavailable settings remain unscored. The balanced subset assigns equal weight to each task.

\section{Results and Analysis}
\label{sec:results}
\label{sec:api}

Table~\ref{tab:main} compares models on the same full question set. The analysis examines visual gain, Earth-normal errors, and differences among the five tasks. Section~4.1 analyzes differences in visual gain, frame order, and rendering. Section~4.2 analyzes Earth-normal errors when the observed dynamics change. Section~4.3 analyzes how the five tasks contribute to the full-set score

\begin{table}[!t]
\centering\small
\setlength{\tabcolsep}{3.1pt}
\renewcommand{\arraystretch}{1.00}
\caption{\textbf{Full-set results.} Score (\%) on the full question set. Gain is the Simulator ordered score minus the Text only score. Models are grouped by developer. Bold marks column maxima within each output token limit, computed before rounding. Missing evaluations remain unscored; failed responses within completed evaluations receive zero credit.}
\label{tab:main}
\label{tab:full_results}
\begin{tabular}{@{}lcrrrrrr@{}}
\toprule
Model & Tokens & Text only & \multicolumn{2}{c}{Simulator} & \multicolumn{2}{c}{Translated} & Gain\\
\cmidrule(lr){4-5}\cmidrule(lr){6-7}
 & & & Ordered & Shuffled & Ordered & Shuffled & (pp)\\
\midrule
\addlinespace[2pt]
\multicolumn{8}{@{}l}{\textbf{Alibaba}}\\[-1pt]
\raisebox{-0.1em}{\includegraphics[width=0.95em,height=0.95em,keepaspectratio]{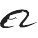}}\hspace{0.3em}Qwen3-VL-32B & 256 & 33.41 & \textbf{47.54} & \textbf{45.35} & \textbf{42.68} & \textbf{42.50} & \textbf{+14.13}\\
\raisebox{-0.1em}{\includegraphics[width=0.95em,height=0.95em,keepaspectratio]{figures/vendor_icons/alibaba.pdf}}\hspace{0.3em}Qwen3-VL-30B-A3B & 256 & 30.76 & 43.87 & 41.12 & 40.03 & 39.12 & +13.11\\
\raisebox{-0.1em}{\includegraphics[width=0.95em,height=0.95em,keepaspectratio]{figures/vendor_icons/alibaba.pdf}}\hspace{0.3em}Qwen3-VL-8B & 256 & 35.67 & 45.76 & 42.75 & 40.35 & 40.15 & +10.09\\
\raisebox{-0.1em}{\includegraphics[width=0.95em,height=0.95em,keepaspectratio]{figures/vendor_icons/alibaba.pdf}}\hspace{0.3em}Qwen3-VL-4B & 256 & 32.20 & 43.26 & 41.25 & 39.21 & 39.19 & +11.07\\
\raisebox{-0.1em}{\includegraphics[width=0.95em,height=0.95em,keepaspectratio]{figures/vendor_icons/alibaba.pdf}}\hspace{0.3em}Qwen3-VL-2B & 256 & 30.79 & 35.55 & 35.05 & 33.35 & 33.48 & +4.75\\
\raisebox{-0.1em}{\includegraphics[width=0.95em,height=0.95em,keepaspectratio]{figures/vendor_icons/alibaba.pdf}}\hspace{0.3em}Qwen2.5-VL-32B & 256 & 33.98 & 39.67 & 40.61 & 34.86 & 34.88 & +5.70\\
\raisebox{-0.1em}{\includegraphics[width=0.95em,height=0.95em,keepaspectratio]{figures/vendor_icons/alibaba.pdf}}\hspace{0.3em}Qwen2.5-VL-7B & 256 & 34.29 & 39.84 & 39.07 & 36.79 & 37.26 & +5.56\\
\raisebox{-0.1em}{\includegraphics[width=0.95em,height=0.95em,keepaspectratio]{figures/vendor_icons/alibaba.pdf}}\hspace{0.3em}Qwen2.5-VL-3B & 256 & 30.93 & 27.68 & 27.26 & 29.55 & 29.32 & -3.25\\
\raisebox{-0.1em}{\includegraphics[width=0.95em,height=0.95em,keepaspectratio]{figures/vendor_icons/alibaba.pdf}}\hspace{0.3em}Qwen2-VL-7B & 256 & 32.13 & 32.56 & 32.53 & 31.29 & 31.47 & +0.42\\
\raisebox{-0.1em}{\includegraphics[width=0.95em,height=0.95em,keepaspectratio]{figures/vendor_icons/alibaba.pdf}}\hspace{0.3em}Qwen2-VL-2B & 256 & 27.56 & 29.48 & 29.59 & 28.91 & 29.08 & +1.93\\
\raisebox{-0.1em}{\includegraphics[width=0.95em,height=0.95em,keepaspectratio]{figures/vendor_icons/alibaba.pdf}}\hspace{0.3em}Qwen3.8-27B & 256 & 36.48 & 43.15 & 41.41 & 40.61 & 40.10 & +6.66\\
\addlinespace[2pt]
\multicolumn{8}{@{}l}{\textbf{Google}}\\[-1pt]
\raisebox{-0.1em}{\includegraphics[width=0.95em,height=0.95em,keepaspectratio]{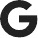}}\hspace{0.3em}Gemma4-12B & 256 & 30.97 & 38.21 & 38.27 & 35.48 & 36.19 & +7.24\\
\addlinespace[2pt]
\multicolumn{8}{@{}l}{\textbf{OpenAI}}\\[-1pt]
\raisebox{-0.1em}{\includegraphics[width=0.95em,height=0.95em,keepaspectratio]{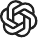}}\hspace{0.3em}GPT-5.4 Mini & 16k & 33.20 & 38.08 & 36.50 & 34.85 & 33.81 & +4.89\\
\raisebox{-0.1em}{\includegraphics[width=0.95em,height=0.95em,keepaspectratio]{figures/vendor_icons/openai.pdf}}\hspace{0.3em}GPT-5.6 Sol & 16k & \textbf{36.65} & \textbf{55.90} & \textbf{43.75} & \textbf{47.84} & \textbf{40.94} & \textbf{+19.25}\\
\raisebox{-0.1em}{\includegraphics[width=0.95em,height=0.95em,keepaspectratio]{figures/vendor_icons/openai.pdf}}\hspace{0.3em}GPT-5.6 Luna & 16k & 33.12 & 44.84 & 35.38 & 38.56 & 34.98 & +11.72\\
\addlinespace[2pt]

\multicolumn{8}{@{}l}{\textbf{Shanghai AI Laboratory}}\\[-1pt]
\raisebox{-0.1em}{\includegraphics[width=0.95em,height=0.95em,keepaspectratio]{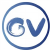}}\hspace{0.3em}InternVL3.5-8B & 256 & 32.15 & 33.42 & 33.11 & 29.15 & 29.25 & +1.27\\
\raisebox{-0.1em}{\includegraphics[width=0.95em,height=0.95em,keepaspectratio]{figures/vendor_icons/shanghai_ailab.pdf}}\hspace{0.3em}InternVL3-8B & 256 & \textbf{38.28} & 43.19 & 39.86 & 38.22 & 37.95 & +4.91\\
\bottomrule
\end{tabular}
\end{table}

\subsection{Overall Results}

\begin{figure}[t]
\centering
\includegraphics[width=\linewidth]{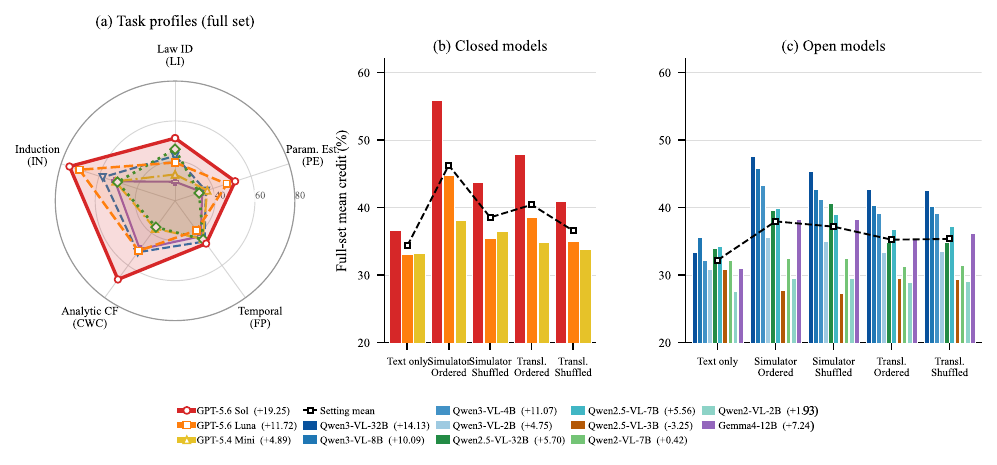}

\caption{\textbf{Task scores across the five input settings.} (a)~Full-set task scores in Simulator ordered for the models in the radar legend; farther from the center means a higher score. (b,c)~Full-set scores (Table~\ref{tab:main}) for the closed and open models with complete five-setting rows. Parentheses in the bar legend give visual gain, the Simulator ordered score minus the Text only score. Dashed lines show each panel's per-setting mean. Family scaling appears in Figure~\ref{fig:scale}.}
\label{fig:setup-sensitivity}
\end{figure}

\textbf{Visual evidence helps most models, and the size of the gain varies widely.}
As shown in Table~\ref{tab:main} and Figure~~\ref{fig:setup-sensitivity}, Qwen2-VL-7B gains 0.42\%, while GPT-5.6 Sol gains 19.25\%, a difference of 18.83\%. Under the smaller output token limit, Qwen3-VL-32B has the largest gain, 14.13\%. The same observations therefore produce markedly different benefits in the full-set score. The subsections below examine whether those answers follow the observed dynamics.

\textbf{Larger models do not improve uniformly across model families.}

As Table~\ref{tab:main} shows, Qwen3-VL-32B leads its family in Simulator ordered setup, by 1.78\% over Qwen3-VL-8B and by 11.99\% over Qwen3-VL-2B. Qwen2.5-VL-32B trails Qwen2.5-VL-7B by 0.17\%. Qwen2.5-VL-3B loses 3.25\% when frames are added. Model size alone does not predict the benefit of visual input.

\textbf{Models respond differently to order and rendering.}

As shown in Figure~\ref{fig:setup-sensitivity} and Figure~\ref{fig:contrasts_full}, the two comparisons do not move together. In Table~2, shuffling simulator frames lowers GPT-5.6 Sol by 12.15\% and Qwen2-VL-7B by 0.03\%. Simulator ordered exceeds Translated ordered for most models by 8.06\% for GPT-5.6 Sol. Qwen2.5-VL-3B reverses the rendering comparison: Translated ordered leads by 1.87\%. These differences separate a benefit from chronological order from a preference for one rendering.

\subsection{Familiar Physics}

\begin{figure}[t]
\centering
\includegraphics[width=\linewidth]{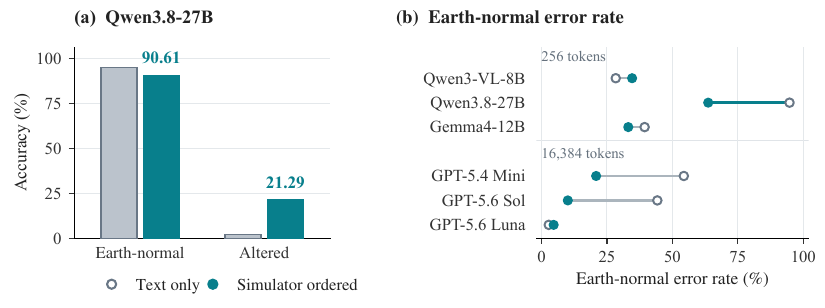}
\caption{\textbf{Familiar physical expectations persist.} Left shows law-identification accuracy on Earth-normal and altered scenes. Right shows the Earth-normal error rate on altered questions. Open and filled markers denote Text only and Simulator ordered. Table~\ref{tab:normal_bias} reports these comparisons.}
\label{fig:normal_bias}
\end{figure}

\textbf{Strong performance on Earth-normal scenes hides errors on altered dynamics.}
Qwen3.8-27B identifies the governing law much more accurately on Earth-normal questions than on altered questions (Figure~\ref{fig:normal_bias}). Simulator ordered substantially improves altered-scene accuracy over a weak Text only result, but the gap between the two scene types remains large. The gain from frames therefore coexists with frequent errors when the observed dynamics change.

\textbf{The Earth-normal answer persists after option positions change.}
Frames reduce Qwen3.8-27B's Earth-normal error rate, yet the Earth-normal answer still dominates its responses to altered questions. The model also repeats that answer after it moves to another option position, indicating a stable semantic preference. Across the diagnostic models, frames can either weaken or strengthen this preference (Table~\ref{tab:normal_bias}).

\subsection{Task-Level Behavior}

\begin{figure}[!htbp]
\centering
\includegraphics[width=0.85\linewidth]{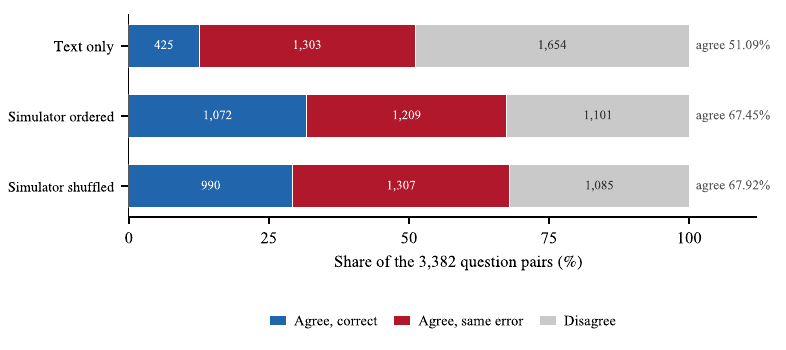}
\vspace{-15pt}
\caption{\textbf{Agreement includes repeated errors.} Each bar separates correct agreement, agreement on the same error, and disagreement for the same law-identification question pairs. Simulator ordered increases correct agreement and reduces wrong agreement, but wrong agreement remains the larger agreeing group. Table~\ref{tab:optionpairs} reports the counts.}
\label{fig:consistency}
\end{figure}

\textbf{Stable answers can repeat the same physical error.}
For Qwen3-VL-8B, Simulator ordered produces more semantic agreement than Text only, yet repeated wrong answers remain more common than correct agreement. Simulator shuffled produces slightly more agreement still, while lowering law-identification accuracy. Agreement therefore measures answer stability. Its correct and wrong groups distinguish stable correct judgments from repeated mistakes (Table~\ref{tab:optionpairs}; Figure~\ref{fig:consistency}). 

\textbf{Matching a physical change does not mean matching its amount.}
Different amount questions keep the same factor or factor combination while changing its amount category. In Simulator ordered, GPT-5.6 Sol's strong overall law-induction result combines high accuracy on Match with poor accuracy on Different amount. GPT-5.4 Mini shows the opposite pattern, rejecting most nonmatching queries but also rejecting most matching ones. The group scores in Table~\ref{tab:induction_subtypes} distinguish matching both factors and amounts from a tendency to answer Yes or No.

\textbf{Frames improve law identification more consistently than analytic counterfactual comparison.}
Every model with paired evaluations improves on law identification, and most improve on temporal reasoning (Figure~\ref{fig:task_gains}). Analytic counterfactual comparison splits evenly between gains and losses. Parameter estimation also improves less consistently than law identification. Qwen3-VL-8B benefits on law identification while losing counterfactual accuracy. Gemma4-12B instead benefits most on parameter estimation. The overall visual gain thus combines different changes in the abilities measured by the five tasks.

\textbf{An almost unchanged overall score can hide large changes in individual tasks.}
Qwen2-VL-7B improves on law identification and temporal reasoning but loses on analytic counterfactual comparison. These opposing task changes largely offset in the full-set score (Table~\ref{tab:all_full_tasks}). A small aggregate gain can therefore coexist with substantial changes in individual tasks.

\begin{figure}[H]
\centering
\includegraphics[width=\linewidth]{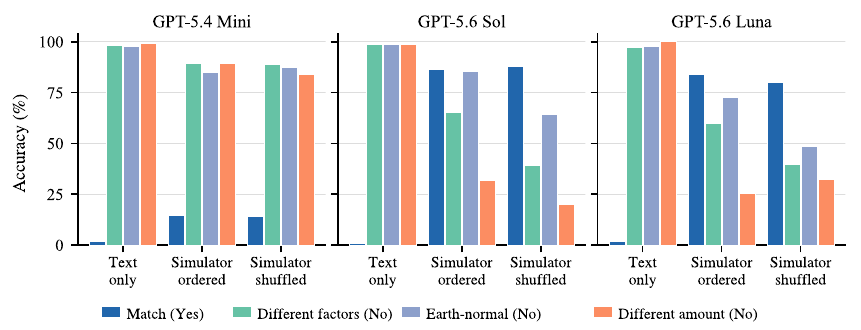}
\caption{\textbf{Accuracy on four kinds of law-induction questions.} Bars report the percentage correct within each group. Match requires the query to share both references' generator category, so the answer is Yes. Different factors, Earth-normal, and Different amount require No. Each model is compared in Text only, Simulator ordered, and Simulator shuffled. Sol and Luna recognize matching questions well with ordered frames but perform much worse when the amount differs (Table~\ref{tab:induction_subtypes}).}
\label{fig:induction-strata}
\vspace{-15pt}
\end{figure}
\begin{figure}[H]
\centering
\includegraphics[width=0.90\linewidth]{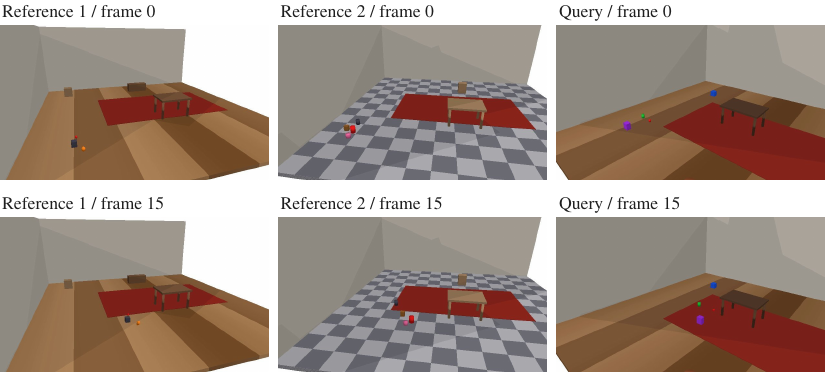}
\caption{\textbf{Matching the factor is insufficient.} This recorded Law induction question asks whether the query shares the reference videos' generator category. The reference answer is No. The query keeps air dissipation but changes its amount category, placing it in the Different amount group. Shown frames are excerpts from the evaluated simulator videos.}
\label{fig:induction_example}
\vspace{-10pt}
\end{figure}

\section{Conclusion}
\label{sec:conclusion}
\benchmark{} evaluates whether physical judgments follow observed dynamics that differ from familiar expectations. Its pipeline produces altered physical scenes, reference answers, and matched questions across frame availability, order, and rendering. Full-set results reveal persistent Earth-normal answer preferences, uneven accuracy across amount categories, and consistent physical errors. Measuring answer quality alongside visual gain, order effect, rendering gap, and task-specific errors characterizes how video-language models use physical evidence.

\FloatBarrier
\newpage
\section*{Ethics Statement}
The benchmark uses synthetic physical scenes without identifiable people. Human annotations concern question usability and contain no demographic information. Simulator and translated videos are identified separately to preserve their provenance.

\section*{Reproducibility Statement}
Sections~\ref{sec:dataset} and~\ref{sec:method} describe benchmark construction and evaluation. The appendix specifies reference answers, model inputs, and human review (Sections~\ref{app:labels}, \ref{app:observations}). Reproduction scripts, selected questions and reference answers, sampled frames, audited score ledgers, figure inputs, and anonymous review labels are released separately from this preprint. These materials support rebuilding the reported tables and figures and recounting the human judgments.

\bibliographystyle{iclr2027_conference}
\bibliography{references}
\par\bigskip
\appendix
\raggedbottom
\setlength{\parskip}{4pt}
\setlength{\textfloatsep}{12pt plus 2pt minus 2pt}
\setlength{\floatsep}{10pt plus 2pt minus 2pt}
\setlength{\intextsep}{12pt plus 2pt minus 2pt}
\Needspace{18\baselineskip}
\section{Rendering Example}
\label{app:rendering}
\begin{figure}[H]
\centering
\includegraphics[width=0.99\linewidth]{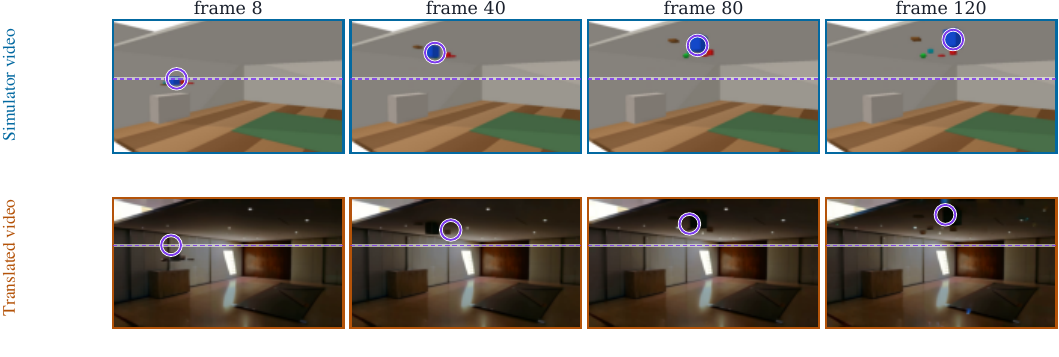}
\caption{\textbf{Paired simulator and translated videos.} An inverted-gravity scene in the simulator video (top) and translated video (bottom). Circles and guides show motion centroids in the displayed source frames. Evaluation uses the common frame sample without these overlays.}
\label{fig:rendering}
\end{figure}

\section{Reference Answers}
\label{app:labels}

\paragraph{Law identification.}
Each simulator video yields two four-option questions, a deterministic permutation and a one-position rotation. For an altered two-factor scene, one of its altered factors is selected as the correct option, and the other altered factor is excluded from the distractors. Each question therefore has one correct listed factor. For an Earth-normal video, the correct option states that the physics matches the reference. The two arrangements form a paired observation of the same video.

\paragraph{Temporal reasoning.}
The generator selects an early moving state and compares the protagonist's vertical position after a fixed time interval. A displacement above $0.25$ m is labelled higher, below $-0.25$ m lower, and absolute displacement at most $0.10$ m approximately unchanged. The intervening ambiguous bands are omitted. Although the option set includes ``out of view,'' no generated reference answer selects that option. The evaluated frames span the full video, so the target measures temporal comparison within a video.

\paragraph{Analytic counterfactual comparison.}
We construct labels by comparing recorded motion with task-specific, idealized Earth-normal references. Height questions use a floor-bounded ballistic reference derived from the recorded initial state. Distance questions use a stopping-distance reference or a predefined rule for eligible dissipation scenes. Earth-normal scenes receive the approximately unchanged label by construction; altered cases that fall within the decision margin or outside the supported rules are excluded. These targets describe comparisons with specified analytic references, rather than outcomes from a matched Earth-normal simulation.

\section{Additional Results}
\label{app:additional}
Table~\ref{tab:main} combines our full-set evaluation results. Every reported cell covers the complete question set. Previously recorded results take priority, followed by complete evaluations with per-question records. This fixed priority is independent of score. Duplicate evaluations remain in the evidence archive.

\subsection{All Full-Set Task Results}
\begingroup\scriptsize\setlength{\tabcolsep}{3pt}
\begin{longtable}{@{}llrrrrrr@{}}
\caption{Full-set task scores (\%). LI is law identification, TR temporal reasoning, AC analytic counterfactual comparison, PE parameter estimation, and IN law induction. Settings are Text only, Simulator ordered, Simulator shuffled, Translated ordered, and Translated shuffled. Macro gives equal weight to each task.}\label{tab:all_full_tasks}\label{tab:alltypes}\label{tab:api_tasks}\\
\toprule
Model & Setting & LI & TR & AC & PE & IN & Macro\\
\midrule
\endfirsthead
\toprule
Model & Setting & LI & TR & AC & PE & IN & Macro\\
\midrule
\endhead
\bottomrule
\endfoot
Qwen3-VL-32B & Text only & 25.78 & 33.63 & 46.45 & 25.85 & 50.00 & 36.34\\
 & Simulator ordered & 45.40 & 51.39 & 51.31 & 36.79 & 51.47 & 47.27\\
 & Simulator shuffled & 44.40 & 42.79 & 51.86 & 36.02 & 50.67 & 45.15\\
 & Translated ordered & 41.40 & 34.85 & 52.56 & 36.83 & 51.20 & 43.37\\
 & Translated shuffled & 40.18 & 37.04 & 53.59 & 34.02 & 50.87 & 43.14\\
\addlinespace[2pt]
Qwen3-VL-30B-A3B & Text only & 21.58 & 35.35 & 40.11 & 25.85 & 50.00 & 34.58\\
 & Simulator ordered & 42.86 & 53.70 & 38.71 & 28.66 & 50.40 & 42.87\\
 & Simulator shuffled & 41.79 & 41.63 & 39.52 & 28.62 & 50.20 & 40.35\\
\addlinespace[2pt]
Qwen3-VL-8B & Text only & 25.61 & 29.17 & 56.69 & 43.78 & 49.47 & 40.94\\
 & Simulator ordered & 42.58 & 45.17 & 51.86 & 36.22 & 58.09 & 46.78\\
 & Simulator shuffled & 39.33 & 37.50 & 53.33 & 34.47 & 56.48 & 44.22\\
 & Translated ordered & 36.28 & 33.33 & 53.04 & 32.93 & 56.02 & 42.32\\
 & Translated shuffled & 35.94 & 33.47 & 52.97 & 33.37 & 55.01 & 42.15\\
\addlinespace[2pt]
Qwen3-VL-4B & Text only & 22.99 & 30.52 & 50.06 & 25.85 & 50.00 & 35.88\\
 & Simulator ordered & 39.86 & 47.02 & 46.00 & 38.50 & 50.00 & 44.28\\
 & Simulator shuffled & 37.30 & 44.61 & 44.53 & 36.63 & 50.13 & 42.64\\
 & Translated ordered & 33.29 & 42.69 & 45.93 & 35.08 & 50.13 & 41.43\\
 & Translated shuffled & 33.55 & 42.26 & 45.52 & 35.53 & 50.07 & 41.39\\
\addlinespace[2pt]
Qwen3-VL-2B & Text only & 22.92 & 26.82 & 35.76 & 50.20 & 49.47 & 37.03\\
 & Simulator ordered & 26.39 & 39.02 & 43.06 & 39.72 & 52.87 & 40.21\\
 & Simulator shuffled & 26.08 & 34.49 & 46.70 & 38.01 & 53.14 & 39.68\\
 & Translated ordered & 25.47 & 32.34 & 41.62 & 39.72 & 50.80 & 37.99\\
 & Translated shuffled & 25.31 & 32.04 & 44.09 & 37.97 & 50.33 & 37.95\\
\addlinespace[2pt]
Qwen2.5-VL-32B & Text only & 24.26 & 28.80 & 54.36 & 35.65 & 50.00 & 38.62\\
 & Simulator ordered & 34.05 & 44.15 & 44.05 & 27.72 & 57.95 & 41.58\\
 & Simulator shuffled & 34.79 & 44.74 & 45.75 & 30.45 & 57.62 & 42.67\\
 & Translated ordered & 27.60 & 35.91 & 40.85 & 33.94 & 55.41 & 38.74\\
 & Translated shuffled & 27.87 & 35.62 & 40.11 & 35.69 & 54.95 & 38.85\\
\addlinespace[2pt]
Qwen2.5-VL-7B & Text only & 28.28 & 33.20 & 44.13 & 29.15 & 50.00 & 36.95\\
 & Simulator ordered & 37.12 & 37.63 & 43.57 & 39.63 & 50.00 & 41.59\\
 & Simulator shuffled & 36.64 & 37.20 & 40.63 & 40.37 & 50.00 & 40.97\\
 & Translated ordered & 36.22 & 24.77 & 42.65 & 40.49 & 50.00 & 38.83\\
 & Translated shuffled & 35.35 & 28.41 & 44.05 & 39.07 & 50.00 & 39.37\\
\addlinespace[2pt]
Qwen2.5-VL-3B & Text only & 24.76 & 26.92 & 42.58 & 25.85 & 50.00 & 34.02\\
 & Simulator ordered & 27.10 & 31.15 & 34.00 & 39.02 & 2.54 & 26.76\\
 & Simulator shuffled & 26.61 & 30.39 & 33.52 & 38.94 & 2.94 & 26.48\\
 & Translated ordered & 26.43 & 31.35 & 36.91 & 41.30 & 16.98 & 30.59\\
 & Translated shuffled & 25.98 & 31.22 & 36.72 & 41.87 & 16.84 & 30.53\\
\addlinespace[2pt]
Qwen2-VL-7B & Text only & 24.26 & 34.03 & 35.54 & 42.15 & 49.47 & 37.09\\
 & Simulator ordered & 28.33 & 38.59 & 20.92 & 40.81 & 53.81 & 36.49\\
 & Simulator shuffled & 28.61 & 35.09 & 24.13 & 40.85 & 53.54 & 36.44\\
 & Translated ordered & 26.89 & 33.00 & 22.50 & 45.81 & 51.67 & 35.98\\
 & Translated shuffled & 26.74 & 32.94 & 23.46 & 45.98 & 52.47 & 36.32\\
\addlinespace[2pt]
Qwen2-VL-2B & Text only & 23.20 & 26.55 & 20.59 & 42.07 & 50.00 & 32.48\\
 & Simulator ordered & 24.36 & 31.58 & 24.90 & 39.02 & 48.86 & 33.75\\
 & Simulator shuffled & 24.08 & 31.75 & 26.00 & 38.98 & 48.86 & 33.94\\
 & Translated ordered & 23.76 & 28.94 & 24.71 & 39.11 & 51.34 & 33.57\\
 & Translated shuffled & 23.86 & 29.27 & 24.49 & 39.02 & 52.47 & 33.82\\
\addlinespace[2pt]
Qwen3.8-27B & Text only & 34.92 & 47.72 & 21.55 & 33.98 & 50.00 & 37.63\\
 & Simulator ordered & 45.85 & 43.98 & 36.28 & 32.68 & 50.33 & 41.82\\
 & Simulator shuffled & 43.33 & 40.91 & 35.99 & 33.50 & 50.07 & 40.76\\
 & Translated ordered & 42.87 & 38.76 & 36.94 & 29.31 & 50.07 & 39.59\\
 & Translated shuffled & 41.11 & 38.66 & 37.50 & 31.79 & 50.00 & 39.81\\
\addlinespace[2pt]
Gemma4-12B & Text only & 23.86 & 33.10 & 45.49 & 9.67 & 50.00 & 32.43\\
 & Simulator ordered & 29.55 & 41.96 & 50.02 & 34.27 & 51.60 & 41.48\\
 & Simulator shuffled & 30.19 & 39.58 & 49.87 & 39.19 & 50.33 & 41.83\\
 & Translated ordered & 26.91 & 37.83 & 47.55 & 29.84 & 52.21 & 38.87\\
 & Translated shuffled & 28.12 & 37.40 & 47.85 & 31.79 & 52.74 & 39.58\\
\addlinespace[2pt]
GPT-5.4 Mini & Text only & 28.73 & 26.49 & 39.52 & 39.96 & 49.93 & 36.93\\
 & Simulator ordered & 33.22 & 43.62 & 37.20 & 36.95 & 51.40 & 40.48\\
 & Simulator shuffled & 34.57 & 31.71 & 38.60 & 36.83 & 50.80 & 38.50\\
 & Translated ordered & 31.45 & 33.27 & 38.82 & 27.89 & 51.94 & 36.67\\
 & Translated shuffled & 31.71 & 26.72 & 39.56 & 30.12 & 50.20 & 35.66\\
\addlinespace[2pt]
GPT-5.6 Sol & Text only & 27.26 & 32.51 & 51.86 & 49.02 & 49.73 & 42.08\\
 & Simulator ordered & 51.39 & 46.49 & 68.69 & 51.63 & 75.60 & 58.76\\
 & Simulator shuffled & 38.84 & 34.29 & 53.74 & 45.28 & 65.71 & 47.57\\
 & Translated ordered & 41.88 & 40.31 & 57.27 & 46.95 & 73.60 & 52.00\\
 & Translated shuffled & 35.98 & 32.04 & 51.20 & 38.54 & 64.71 & 44.49\\
\addlinespace[2pt]
GPT-5.6 Luna & Text only & 22.37 & 28.74 & 49.10 & 47.28 & 50.00 & 39.50\\
 & Simulator ordered & 39.19 & 38.29 & 51.05 & 47.32 & 70.32 & 49.23\\
 & Simulator shuffled & 29.76 & 25.43 & 44.13 & 40.77 & 60.63 & 40.14\\
 & Translated ordered & 33.19 & 30.85 & 47.11 & 41.79 & 60.29 & 42.65\\
 & Translated shuffled & 29.46 & 27.31 & 43.43 & 36.71 & 58.62 & 39.11\\
\addlinespace[2pt]
MiniCPM-V-4.5 & Text only & 26.03 & 34.19 & 55.62 & 40.57 & 50.53 & 41.39\\
\addlinespace[2pt]
InternVL3.5-8B & Text only & 22.16 & 29.50 & 58.86 & 12.89 & 50.00 & 34.68\\
\addlinespace[2pt]
InternVL3-8B & Text only & 32.30 & 37.70 & 49.28 & 34.07 & 50.00 & 40.67\\
 & Simulator ordered & 39.06 & 37.04 & 57.24 & 40.85 & 50.74 & 44.98\\
\addlinespace[2pt]
\end{longtable}

\endgroup
\begin{table}[!htbp]
\centering\small
\caption{\textbf{Qwen3VL-8B option-arrangement consistency.} Correct agreement, wrong agreement, valid disagreement, and invalid pairs form an exhaustive partition of the full set of question pairs. Agree is the percentage of pairs whose valid responses give identical answer text, computed from Correct plus Wrong before rounding. High agreement need not imply high accuracy.}
\label{tab:optionpairs}
\begin{tabular}{@{}lrrrrr@{}}
\toprule
Setting & Agree (\%) & Correct ($n$) & Wrong ($n$) & Disagree ($n$) & Invalid ($n$)\\
\midrule
Text only & 51.09 & 425 & 1,303 & 1,654 & 0\\
Simulator ordered & 67.45 & 1,072 & 1,209 & 1,101 & 0\\
Simulator shuffled & 67.92 & 990 & 1,307 & 1,085 & 0\\
\bottomrule
\end{tabular}
\end{table}

\section{Additional Evaluation Settings}
\label{app:additional_settings}

\paragraph{Balanced subset.}
The balanced subset gives each task equal weight. It retains paired option arrangements for law identification and balances Match and nonmatching questions for law induction. Table~\ref{tab:subset} reports every input setting for each completed model. Compatible saved responses are reused, and new evaluations cover only missing identifiers.
\begin{table}[!htbp]
\centering\small
\setlength{\tabcolsep}{3.0pt}
\renewcommand{\arraystretch}{1.00}
\caption{\textbf{Balanced-subset results.} Task-balanced score (\%) on the same questions across input settings. Gain is the Simulator ordered score minus the Text only score. Tokens denotes the output limit. Bold marks column maxima within each output-limit group, using unrounded scores. Failed responses receive zero credit. Previously saved compatible answers are rescored on these exact question identifiers. Model names identify the evaluated systems. Table~\ref{tab:main} reports the full-set comparison. SmolVLM2 uses its fixed native video preprocessing while receiving the same external frames.}
\label{tab:subset}
\begin{tabular}{@{}lcrrrrrr@{}}
\toprule
Model & Tokens & Text only & \multicolumn{2}{c}{Simulator} & \multicolumn{2}{c}{Translated} & Gain\\
\cmidrule(lr){4-5}\cmidrule(lr){6-7}
 & & & Ordered & Shuffled & Ordered & Shuffled & (pp)\\
\midrule
\addlinespace[2pt]
\multicolumn{8}{@{}l}{\textbf{Alibaba}}\\[-1pt]
\raisebox{-0.10em}{\includegraphics[width=0.95em,height=0.95em,keepaspectratio]{figures/vendor_icons/alibaba.pdf}}\hspace{0.3em}Qwen3-VL-32B & 256 & 34.90 & \textbf{45.95} & \textbf{45.20} & 40.45 & \textbf{42.25} & \textbf{+11.05}\\
\raisebox{-0.10em}{\includegraphics[width=0.95em,height=0.95em,keepaspectratio]{figures/vendor_icons/alibaba.pdf}}\hspace{0.3em}Qwen3-VL-30B-A3B & 256 & 32.30 & 42.85 & 40.45 & 39.95 & 38.45 & +10.55\\
\raisebox{-0.10em}{\includegraphics[width=0.95em,height=0.95em,keepaspectratio]{figures/vendor_icons/alibaba.pdf}}\hspace{0.3em}Qwen3-VL-8B & 256 & 38.65 & 45.80 & 42.60 & 40.10 & 40.35 & +7.15\\
\raisebox{-0.10em}{\includegraphics[width=0.95em,height=0.95em,keepaspectratio]{figures/vendor_icons/alibaba.pdf}}\hspace{0.3em}Qwen3-VL-4B & 256 & 35.00 & 44.15 & 43.10 & 40.15 & 40.15 & +9.15\\
\raisebox{-0.10em}{\includegraphics[width=0.95em,height=0.95em,keepaspectratio]{figures/vendor_icons/alibaba.pdf}}\hspace{0.3em}Qwen3-VL-2B & 256 & 34.35 & 38.25 & 38.10 & 37.40 & 37.40 & +3.90\\
\raisebox{-0.10em}{\includegraphics[width=0.95em,height=0.95em,keepaspectratio]{figures/vendor_icons/alibaba.pdf}}\hspace{0.3em}Qwen2.5-VL-32B & 256 & 37.60 & 39.55 & 39.55 & 38.30 & 37.90 & +1.95\\
\raisebox{-0.10em}{\includegraphics[width=0.95em,height=0.95em,keepaspectratio]{figures/vendor_icons/alibaba.pdf}}\hspace{0.3em}Qwen2.5-VL-7B & 256 & 36.45 & 41.75 & 41.80 & 38.45 & 39.15 & +5.30\\
\raisebox{-0.10em}{\includegraphics[width=0.95em,height=0.95em,keepaspectratio]{figures/vendor_icons/alibaba.pdf}}\hspace{0.3em}Qwen2.5-VL-3B & 256 & 33.20 & 25.80 & 26.55 & 29.90 & 29.90 & -7.40\\
\raisebox{-0.10em}{\includegraphics[width=0.95em,height=0.95em,keepaspectratio]{figures/vendor_icons/alibaba.pdf}}\hspace{0.3em}Qwen2-VL-7B & 256 & 35.60 & 34.70 & 35.75 & 33.45 & 34.65 & -0.90\\
\raisebox{-0.10em}{\includegraphics[width=0.95em,height=0.95em,keepaspectratio]{figures/vendor_icons/alibaba.pdf}}\hspace{0.3em}Qwen2-VL-2B & 256 & 30.70 & 32.75 & 33.15 & 32.85 & 32.75 & +2.05\\
\raisebox{-0.10em}{\includegraphics[width=0.95em,height=0.95em,keepaspectratio]{figures/vendor_icons/alibaba.pdf}}\hspace{0.3em}Qwen3.8-27B & 256 & 37.00 & 42.10 & 40.75 & 40.50 & 40.00 & +5.10\\
\addlinespace[2pt]
\multicolumn{8}{@{}l}{\textbf{Google}}\\[-1pt]
\raisebox{-0.10em}{\includegraphics[width=0.95em,height=0.95em,keepaspectratio]{figures/vendor_icons/google.pdf}}\hspace{0.3em}Gemma4-12B & 256 & 32.00 & 40.70 & 39.35 & 39.60 & 41.10 & +8.70\\
\raisebox{-0.10em}{\includegraphics[width=0.95em,height=0.95em,keepaspectratio]{figures/vendor_icons/google.pdf}}\hspace{0.3em}Gemini 3.8 Flash & 16k & 39.00 & 54.90 & 47.45 & \textbf{53.15} & \textbf{46.00} & +15.90\\
\addlinespace[2pt]
\multicolumn{8}{@{}l}{\textbf{OpenAI}}\\[-1pt]
\raisebox{-0.10em}{\includegraphics[width=0.95em,height=0.95em,keepaspectratio]{figures/vendor_icons/openai.pdf}}\hspace{0.3em}GPT-5.4 Nano & 16k & 33.20 & 34.50 & 35.90 & 33.05 & 33.60 & +1.30\\
\raisebox{-0.10em}{\includegraphics[width=0.95em,height=0.95em,keepaspectratio]{figures/vendor_icons/openai.pdf}}\hspace{0.3em}GPT-5.4 Mini & 16k & 36.10 & 39.05 & 37.50 & 37.35 & 35.95 & +2.95\\
\raisebox{-0.10em}{\includegraphics[width=0.95em,height=0.95em,keepaspectratio]{figures/vendor_icons/openai.pdf}}\hspace{0.3em}GPT-5.6 Sol & 16k & \textbf{42.05} & \textbf{59.00} & \textbf{48.05} & 51.25 & 44.65 & \textbf{+16.95}\\
\raisebox{-0.10em}{\includegraphics[width=0.95em,height=0.95em,keepaspectratio]{figures/vendor_icons/openai.pdf}}\hspace{0.3em}GPT-5.6 Luna & 16k & 38.60 & 48.80 & 40.20 & 40.75 & 37.90 & +10.20\\
\addlinespace[2pt]
\multicolumn{8}{@{}l}{\textbf{OpenBMB}}\\[-1pt]
\raisebox{-0.10em}{\includegraphics[width=0.95em,height=0.95em,keepaspectratio]{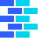}}\hspace{0.3em}MiniCPM-V-4.5 & 256 & \textbf{40.90} & 45.35 & 44.90 & 40.20 & 40.95 & +4.45\\
\addlinespace[2pt]
\multicolumn{8}{@{}l}{\textbf{Shanghai AI Laboratory}}\\[-1pt]
\raisebox{-0.10em}{\includegraphics[width=0.95em,height=0.95em,keepaspectratio]{figures/vendor_icons/shanghai_ailab.pdf}}\hspace{0.3em}InternVL3.5-8B & 256 & 32.65 & 37.20 & 36.90 & 37.60 & 37.30 & +4.55\\
\raisebox{-0.10em}{\includegraphics[width=0.95em,height=0.95em,keepaspectratio]{figures/vendor_icons/shanghai_ailab.pdf}}\hspace{0.3em}InternVL3-8B & 256 & 40.75 & 45.90 & 44.75 & \textbf{42.10} & 41.85 & +5.15\\
\addlinespace[2pt]
\multicolumn{8}{@{}l}{\textbf{Hugging Face}}\\[-1pt]
\raisebox{-0.10em}{\includegraphics[width=0.95em,height=0.95em,keepaspectratio]{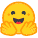}}\hspace{0.3em}SmolVLM2-500M & 256 & 32.80 & 31.15 & 30.50 & 29.45 & 29.90 & -1.65\\
\bottomrule
\end{tabular}
\end{table}

\subsection{Balanced-Subset Results}
\label{app:subset_results}

Table~\ref{tab:subset} compares fully evaluated models on the same balanced subset. Tasks contribute equally to the score, and each row contains every input setting. Paired comparisons retain each model's output token limit.

Adding frames yields substantial gains for some models and losses for others, and restoring chronological order changes scores sharply for some models (e.g., GPT-5.6~Sol) while barely moving others.

\textbf{More visual input does not guarantee a better answer.}
On the balanced subset, Simulator ordered substantially helps Gemini 3.8 Flash but lowers Qwen2-VL-7B's score despite valid answers throughout. Qwen2-VL-7B improves on temporal reasoning while losing more on analytic counterfactual comparison. Visual benefit therefore depends on the task and model (Figure~\ref{fig:setup-sensitivity}).

\textbf{Answer production and physical correctness can fail separately.}
Qwen2.5-VL-3B loses score with frames while producing unparseable answers and performing poorly on law induction. SmolVLM2 also declines and leaves some Simulator ordered questions without a completed response. These unusable outputs receive no credit and lower the reported score.

\begin{figure}[!htbp]
\centering
\includegraphics[width=0.99\linewidth]{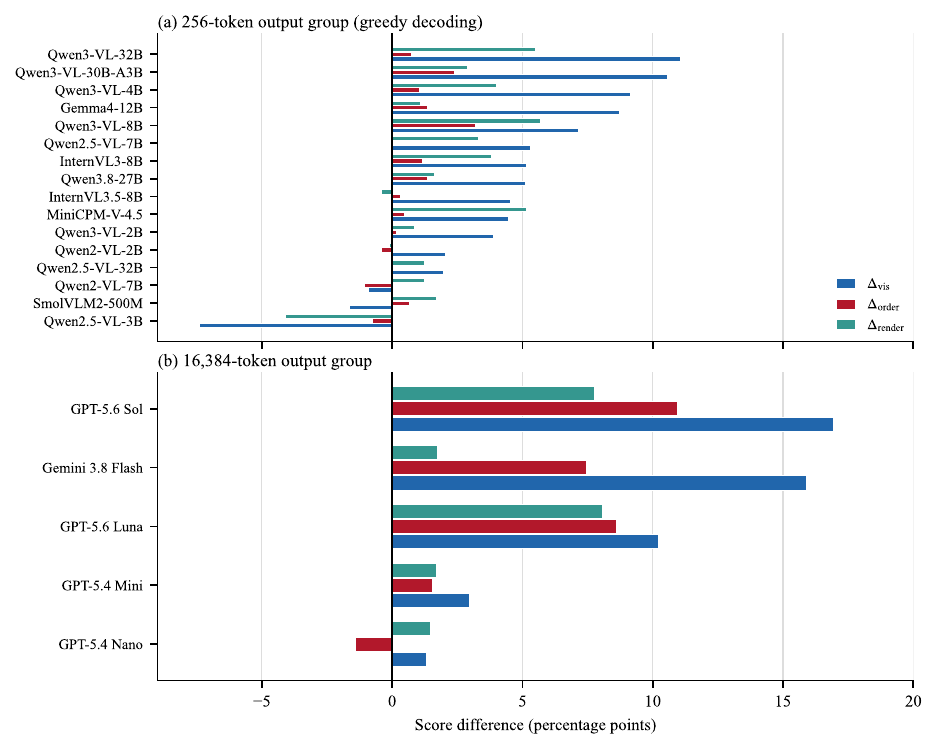}
\caption{\textbf{Paired input-setting comparisons on the balanced subset.} Visual gain adds Simulator ordered frames, the order effect restores chronological presentation, and the rendering gap compares Simulator ordered with Translated ordered (Eq.~\ref{eq:contrasts}; Table~\ref{tab:subset}). Models are sorted by visual gain within each output-budget group. Both visual gain and the order effect vary in sign, showing that frames and chronological order do not improve every model's score.}
\label{fig:contrasts}
\end{figure}

\textbf{Sensitivity to sequence varies more than score alone suggests.}
Restoring chronological order helps GPT-5.6 Sol more than Gemini 3.8 Flash. Models can therefore respond differently to the same input change even under the same output token limit. Figure~\ref{fig:contrasts} compares these effects with visual gain and rendering gap.

\textbf{Changing the rendering can help or hurt.}
MiniCPM-V-4.5 favors Simulator ordered over Translated ordered, while InternVL3.5-8B slightly favors the translated rendering. These matched-question differences combine changes in appearance and question-relevant visual evidence.

\begin{figure}[!htbp]
\centering
\includegraphics[width=0.72\linewidth]{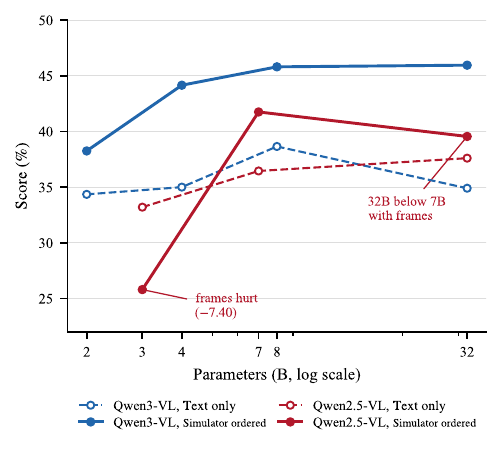}
\caption{\textbf{Scaling does not guarantee greater visual benefit.} Balanced-subset scores versus model size for Qwen3-VL and Qwen2.5-VL in Text only and Simulator ordered (Table~\ref{tab:subset}; Qwen3-VL-30B-A3B is omitted). Qwen2.5-VL-3B loses score with frames, while the family's largest model remains below its next smaller variant in Simulator ordered.}
\label{fig:scale}
\end{figure}

\begin{figure}[!htbp]
\centering
\includegraphics[width=0.82\linewidth]{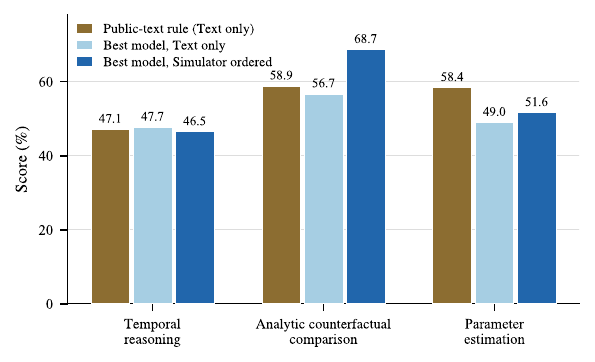}
\caption{\textbf{Public-text rules rival model scores on some tasks.} Rules use question text and options and were selected after inspecting labels (Table~\ref{tab:priors}). Model maxima use the diagnostic group in Figure~\ref{fig:setup-sensitivity}(a), comprising Qwen3-VL-8B, Qwen3.8-27B, Gemma4-12B, GPT-5.4 Mini, GPT-5.6 Sol, and GPT-5.6 Luna. The parameter-estimation rule exceeds every visual-setting score in this group; the temporal rule approaches its best Text only score. Matched frame comparisons help distinguish these answer patterns from visual benefit.}
\label{fig:template}
\end{figure}

\begin{figure}[!htbp]
\centering
\includegraphics[width=\linewidth]{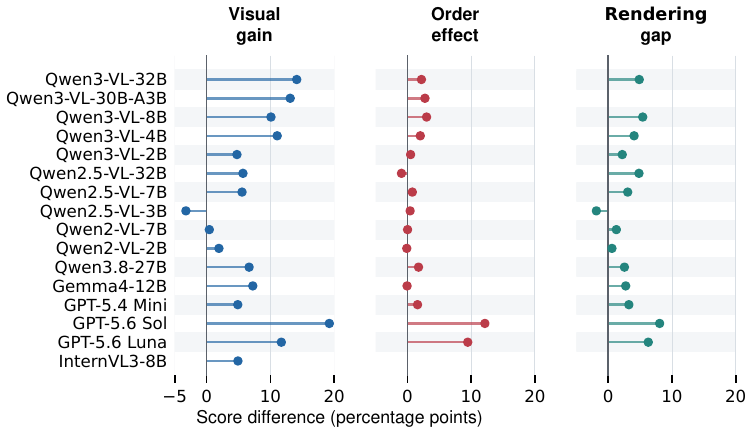}
\caption{\textbf{Sensitivity to supplied evidence.} Differences in full-set score for the same model and questions. Visual gain adds Simulator ordered frames, order effect restores chronological order, and rendering gap compares simulator with translated frames. Positive values favor the first setting in each definition. Only complete pairs are shown.}
\label{fig:contrasts_full}
\end{figure}

\begin{figure}[!htbp]
\centering
\includegraphics[width=\linewidth]{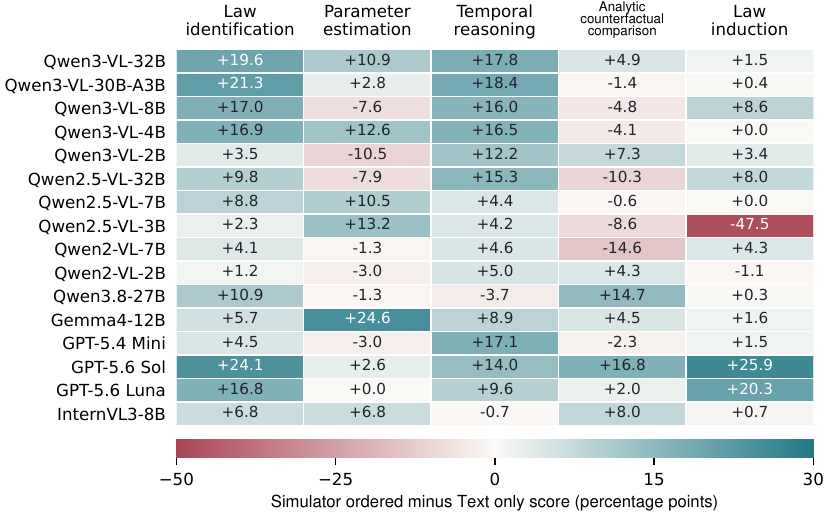}
\caption{\textbf{Visual gains depend on the task.} Simulator ordered score minus Text only score for models with both full-set evaluations. Each column retains every question in its task. Differences are computed before rounding; positive values indicate a benefit from frames. The columns separate law identification, parameter estimation, temporal reasoning, analytic counterfactual comparison, and law induction.}
\label{fig:task_gains}
\end{figure}

\section{Scene Generation}
\label{app:generation}
The generation record covers Earth-normal scenes, individual altered factors, and factor combinations. Category sizes vary, with more candidates in the friction category than in the other individual altered categories. Accepted scene identifiers exactly match the simulator videos referenced by the evaluation questions.

\paragraph{Implementation.}
The stored simulator records specify \texttt{v2\_engine\_stepped}, fixed simulation and rendering rates, a speed limit, and activation of altered parameters at the start of the scene. The implementation uses PyBullet in DIRECT mode and TinyRenderer. It applies horizontal acceleration $(2\omega v_y,-2\omega v_x)$ for the chiral term and velocity-proportional damping for coefficient $k$. Positive $g_z$ points upward.

\paragraph{Scene Templates.}
The accepted records use scene templates associated with their factor or factor combination. For example, gravity-direction scenes include levitation layouts and restitution scenes include staircases. Camera and scene seeds vary across accepted videos, and appearance and layout vary with these templates. For chiral generator categories, the category fixes $s_\omega$, while effective $\omega$ varies by template. Law induction consequently matches the generator category, including its template-dependent coefficient variation.

\paragraph{Quality Checks.}
Scene acceptance uses fit checks and a bounded retry procedure. Most accepted scenes have an eligible component fit; Earth-normal dissipation and chiral scenes can lack one. Rejected candidates include low-contrast renders, floor exits, and a specification-overlap failure. Some accepted records include camera repairs.

The source checks finite trajectories, floor containment with vertical threshold $-0.12$ m, and sufficient protagonist motion. It fits gravity on eligible contact-free segments with error tolerance $0.12\max(|g_z|,1)$ when drag is absent, damping with $0.25\max(k,0.5)$, and altered chiral rates with $0.2\max(\omega,0.1)$. Sliding fits distinguish near-zero friction ($\hat\mu\leq0.03$) from Earth-normal scenes ($\hat\mu\geq0.05$). Restitution requires separated bounces and agreement with the nominal ratio. Undefined $R^2$ values are accepted when the remaining checks pass, while missing eligible segments can fail other checks. Low grayscale contrast also rejects a render. Human review assesses usability with the sampled frames supplied for evaluation.

\section{Model Inputs}
\label{app:observations}
Both prompt builders start with ``You are answering questions about physics in videos. Respond with ONLY the answer.'' Input with frames supplies a video marker followed by its frames; law induction repeats this structure for the reference and query videos. The public question and numbered options follow, ending with an option-number or numeric-answer instruction. Text only retains the same markers and question text but omits frames. Construction metadata and the scoring key remain separate from the model input.

The evaluated frames use the common index schedule
\[
(0,15,29,44,58,73,87,102,116,131,145,160).
\]
Every evaluated video uses this schedule. Prompts present the frames without timestamp, frame-rate, or duration annotations. Shuffling changes their presentation order within each video.

The temporal-reasoning generator selects state times before formatting them in the question. These instants usually fall between sampled frames, and no question includes both queried instants exactly in its sampled observations. Many queries concern early motion. Answering them requires temporal alignment from sparse observations. Parameter estimation similarly relies on scale and timing cues in the frames.

\section{Answer Patterns}
\label{app:diagnostics}
The deterministic rules in Table~\ref{tab:priors} inspect only public question wording and option text. They were selected after inspecting the evaluation labels; reference answers are used only for scoring. They describe answer regularities in the evaluated question set. All task questions remain in each denominator, with numeric half credit unchanged.
\begin{table}[!htbp]
\centering\footnotesize
\caption{Post hoc analysis of answer distributions (score, \%). Rules use public question text and options only, but were selected after inspecting test labels. They are not trained baselines or held-out generalization estimates. Numeric scoring retains half credit.}
\label{tab:priors}
\begin{tabular}{@{}p{2.5cm}p{7.4cm}rr@{}}
\toprule
Task & Public-text prediction rule & Credit / $N$ & Score\\
\midrule
Temporal reasoning & Always same height & 1425 / 3,024 & 47.12\\
Analytic counterfactual comparison & Lower for height; same for total distance & 1598 / 2,715 & 58.86\\
Parameter estimation & Gravity ratio 1.0; upward fraction 0.25; bounce ratio 0.25 & 718.5 / 1,230 & 58.41\\
\bottomrule
\end{tabular}
\end{table}

Analytic counterfactual height questions favor lower or unchanged answers, while distance questions include unchanged, shorter, and longer outcomes. Conditioning a rule on the question template therefore captures more structure than choosing a globally common answer. Fixed numeric predictions also obtain credit on parameter estimation. Table~\ref{tab:priors} shows how these answer patterns can support performance without frames.

\subsection{Task Exclusions}
\label{app:sensitivity}
The full-set tables retain the complete question set. The following analysis recomputes scores after excluding analytic counterfactual comparison, then also temporal reasoning. These exclusions examine how the analytic reference rules and temporal alignment affect the findings. Every remaining question and failure stays in the denominator. Both visual gains and order effects can change.
\begin{table}[!htbp]
\centering\footnotesize
\caption{Post hoc full-set simulator task-exclusion sensitivity. Simulator ordered scores are percentages and contrasts are points calculated from unrounded credits. Every retained question and failure remains included. Output limits differ across model groups.}
\label{tab:sensitivity}
\begin{tabular}{@{}lrrrrrr@{}}
\toprule
Model & \multicolumn{3}{c}{Exclude AC} & \multicolumn{3}{c}{Exclude AC and TR}\\
\midrule
 & \shortstack{Simulator\\ordered} & Visual gain & Order effect & \shortstack{Simulator\\ordered} & Visual gain & Order effect\\
\midrule
Qwen3-VL-8B & 44.43 & +13.33 & +3.98 & 44.20 & +12.48 & +2.80\\
Qwen3.8-27B & 44.64 & +4.91 & +2.05 & 44.85 & +7.67 & +1.73\\
Gemma4-12B & 35.65 & +7.83 & -0.10 & 33.64 & +7.50 & -0.89\\
GPT-5.4 Mini & 38.27 & +6.45 & +2.23 & 36.57 & +3.05 & -0.85\\
GPT-5.6 Sol & 53.12 & +19.77 & +11.54 & 55.24 & +21.61 & +11.33\\
GPT-5.6 Luna & 43.50 & +13.84 & +10.01 & 45.15 & +15.20 & +9.10\\
\bottomrule
\end{tabular}
\end{table}

\section{Earth-normal Answers}
\label{app:normal_answers}
We examine the semantic answers selected on altered law-identification questions. The Earth-normal answer is wrong on these questions and changes position between their option arrangements. Earth-normal error rates and accuracy retain every assigned question, including failures. Table~\ref{tab:normal_bias} compares Text only and Simulator ordered.

\begin{table}[!htbp]
\centering
\small
\setlength{\tabcolsep}{4pt}
\caption{Earth-normal error rates on altered law-identification questions. Text only and Simulator ordered columns give the percentage choosing the Earth-normal answer. Difference is Simulator ordered minus Text only in percentage points. Altered-scene accuracy is the Simulator ordered accuracy on the same questions. Failures remain in every denominator. This diagnostic uses all altered law-identification questions. Model groups retain their original output budgets. All values are rounded independently from integer counts.}
\label{tab:normal_bias}
\begin{tabular}{lrrrr}
\toprule
Model & Text only & Simulator ordered & Difference (pp) & Accuracy \\
\midrule
Qwen3-VL-8B & 28.37 & 34.62 & $+6.25$ & 31.23 \\
Qwen3.8-27B & 94.76 & 63.69 & $-31.07$ & 21.29 \\
Gemma4-12B & 39.40 & 33.15 & $-6.25$ & 26.14 \\
\midrule
GPT-5.4 Mini & 54.37 & 20.88 & $-33.49$ & 33.79 \\
GPT-5.6 Sol & 44.25 & 10.05 & $-34.20$ & 59.07 \\
GPT-5.6 Luna & 2.77 & 4.65 & $+1.88$ & 52.93 \\
\bottomrule
\end{tabular}
\end{table}

Qwen3.8-27B frequently selects the Earth-normal answer across factor groups, including factor combinations. Simulator ordered reduces this error, but the preference persists across option positions and can repeat under both arrangements. These patterns are computed from complete full-set responses.

Some models increase their Earth-normal error rate after receiving frames while also improving altered-scene accuracy. The specific error rate and overall accuracy therefore describe different aspects of their responses.

\section{Law-Induction Results}
\label{app:induction}
Law induction asks whether the query shares the reference videos' generator category. Match requires Yes; Different factors, Earth-normal, and Different amount require No. Reference videos use altered physics even when the query is Earth-normal. The comparison operates within each question, using categories that include template-dependent variation in effective coefficients.

Table~\ref{tab:induction_subtypes} compares the question groups in Text only, Simulator ordered, and Simulator shuffled. Rescoring the saved responses reproduces the reported law-induction accuracies. Sol's strong Simulator ordered result coexists with weak performance on Different amount. In Text only, the models usually answer No, rejecting most nonmatching queries but also most matching ones. Mini retains this pattern with frames. The group scores expose these response preferences in a task with balanced Yes and No reference answers.

\begin{table}[!htbp]
\centering\footnotesize
\caption{Law-induction accuracy by question group (\%). Parentheses show the reference answer. Failed responses receive zero credit. The group scores distinguish category matching from a general tendency to answer No.}
\label{tab:induction_subtypes}
\begin{tabular}{@{}llrrrrr@{}}
\toprule
Model & Setting & \shortstack{Match\\(Yes)} & \shortstack{Different factors\\(No)} & \shortstack{Earth-normal\\(No)} & \shortstack{Different amount\\(No)} & All\\
\midrule
Mini & Text only & 1.60 & 98.01 & 97.93 & 99.35 & 49.93\\
Mini & Simulator ordered & 14.71 & 89.49 & 85.12 & 89.61 & 51.40\\
Mini & Simulator shuffled & 14.17 & 88.92 & 87.60 & 83.77 & 50.80\\
Sol & Text only & 0.67 & 98.86 & 98.76 & 98.70 & 49.73\\
Sol & Simulator ordered & 86.23 & 65.34 & 85.54 & 31.82 & 75.60\\
Sol & Simulator shuffled & 88.10 & 39.20 & 64.05 & 20.13 & 65.71\\
Luna & Text only & 2.01 & 97.44 & 97.52 & 100.00 & 50.00\\
Luna & Simulator ordered & 83.82 & 59.66 & 72.73 & 25.32 & 70.32\\
Luna & Simulator shuffled & 80.21 & 39.49 & 48.76 & 32.47 & 60.63\\
\bottomrule
\end{tabular}
\end{table}

\begingroup\fontsize{9}{9.8}\selectfont\setlength{\parskip}{0pt}
\section{Related Work}
\label{app:related}

\noindent\textbf{Intuitive Physics.}\quad
Simulation-based accounts model human physical scene judgments~\citep{battaglia2013}.
Galileo connects visual property inference to a physics engine~\citep{galileo},
whereas Interaction Networks learn dynamics from object and relation
representations~\citep{interactionnetworks}.
IntPhys tests event plausibility, PLATO learns object-centric physical
concepts, and PHYRE tests action selection~\citep{intphys,plato,phyre}.
Physion and Physion++ examine prediction and online property inference;
ContPhy addresses continuum matter~\citep{physion,physionpp,contphy}.
PhysBench covers broader physical-world questions, while IntPhys~2 and GRASP
extend tests of physical reasoning~\citep{physbench,intphys2,grasp}.
We vary simulated physics and visual input to distinguish familiar knowledge from observed evidence.

\noindent\textbf{Video Understanding.}\quad
CLEVR varies the composition of questions, and CATER isolates compositional
actions and temporal reasoning~\citep{clevr,cater}.
NExT-QA emphasizes causal and temporal questions; Perception Test probes
multimodal skills~\citep{nextqa,perceptiontest}.
MVBench covers diverse temporal tasks, Video-MME spans durations and
modalities, and TempCompass uses conflicting videos with shared static
content~\citep{mvbench,videomme,tempcompass}.
We compare frame orders while retaining identical observations.

\noindent\textbf{Counterfactual Reasoning.}\quad
CLEVRER includes predictive and counterfactual questions~\citep{clevrer}.
CoPhy learns intervention-conditioned trajectories, and Filtered-CoPhy
constrains counterfactual prediction in pixel space~\citep{cophy,filteredcophy}.
Our analytic targets differ from collision-resolved simulation.

\noindent\textbf{Video Generation.}\quad
VBench separates video-quality dimensions; WorldModelBench evaluates
instruction following and physics adherence~\citep{vbench,worldmodelbench}.
VideoPhy evaluates physical commonsense in generated videos, and Physics-IQ compares
generated continuations with physical events~\citep{videophy,physicsiq}.
Physics-IQ Verified documents effects of prompt ambiguity, metrics, and
aggregation~\citep{physicsiqverified}.
Kubric supports annotated scene generation~\citep{kubric}; domain
randomization varies appearance for transfer~\citep{domainrand}; Cosmos
provides world models and conditional generation~\citep{cosmoswfm,cosmostransfer}.
We compare renderings with shared reference answers, including any translation-induced changes.

\noindent\textbf{Evaluation.}\quad
Shortcut-learning studies and fresh ImageNet test sets expose unintended cues and sampling effects~\citep{shortcut,texturebias,imagenetv2}.
VQA~v2 reduces language priors through paired images, while VQA-CP changes answer distributions~\citep{vqav2,vqacp}.
Multiple-choice studies test option-order and identifier effects~\citep{optionorder,mcqrobust}.
CheckList, HELM, and Datasheets for Datasets guide capability evaluation and dataset reporting~\citep{checklist,helm,datasheets}.
Matched input settings distinguish answer quality from the effect of visual evidence.

\endgroup

\clearpage

\section{Examples}

\label{app:examples}

The following questions cover the five tasks. Each shows selected frames from its evaluated input and reproduces the original question, options, and reference answer. The displayed frames retain their original order; video-frame indices identify the selected observations.

\par\vspace{2pt}

\begin{figure}[H]

\centering

\includegraphics[width=\linewidth]{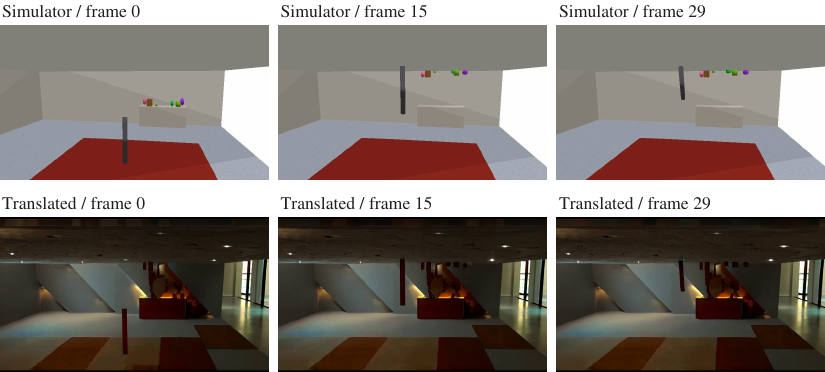}

\par\vspace{3pt}

\begin{minipage}{\linewidth}
\fontsize{9}{10.6}\selectfont\raggedright
\textbf{Question}\quad In this video, which one of the following aspects of physics behaves differently from normal Earth physics?\par
\vspace{2pt}
\textbf{Options}\par
\begin{tabular}{@{}p{0.48\linewidth}@{\hspace{0.04\linewidth}}p{0.48\linewidth}@{}}
(1) None - the physics matches normal Earth physics & (2) The direction of gravity\\
(3) Surface friction & (4) The strength of gravity\\
\end{tabular}\par
\vspace{2pt}
\textbf{Reference answer}\quad 2. The direction of gravity
\end{minipage}

\caption{Law identification. Selected frames from the Simulator and Translated videos.}

\label{fig:example_li}

\end{figure}

\begin{figure}[H]

\centering

\includegraphics[width=\linewidth]{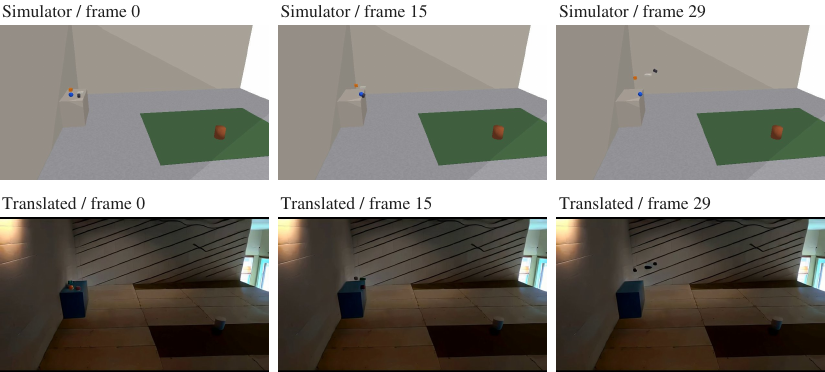}

\par\vspace{3pt}

\begin{minipage}{\linewidth}
\fontsize{9}{10.6}\selectfont\raggedright
\textbf{Question}\quad When an object bounces on the floor in this video, estimate the ratio of a bounce's peak height to the previous bounce's peak height (dimensionless; below 1.0 means bounces shrink, above 1.0 means they grow).\par
\vspace{2pt}
\textbf{Reference answer}\quad 1.69 (dimensionless)
\end{minipage}

\caption{Parameter estimation. A numeric reference answer for the bounce-height ratio.}

\label{fig:example_pe}

\end{figure}

\clearpage

\begin{figure}[H]

\centering

\includegraphics[width=\linewidth]{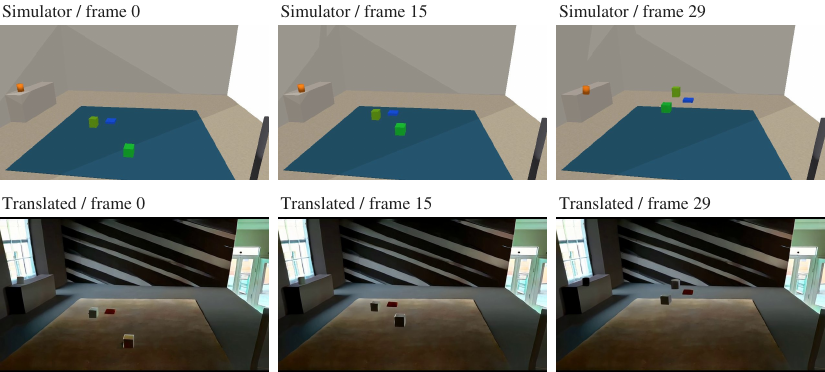}

\par\vspace{3pt}

\begin{minipage}{\linewidth}
\fontsize{9}{10.6}\selectfont\raggedright
\textbf{Question}\quad Consider the position of the moving object at t=0.69 s in the video. At t=1.19 s, where is the object relative to that position?\par
\vspace{2pt}
\textbf{Options}\par
\begin{tabular}{@{}p{0.48\linewidth}@{\hspace{0.04\linewidth}}p{0.48\linewidth}@{}}
(1) At about the same height & (2) It is out of view (off-screen)\\
(3) Noticeably lower & (4) Noticeably higher\\
\end{tabular}\par
\vspace{2pt}
\textbf{Reference answer}\quad 4. Noticeably higher
\end{minipage}

\caption{Temporal reasoning. The two query times fall between the sampled frames.}

\label{fig:example_tr}

\end{figure}

\begin{figure}[H]

\centering

\includegraphics[width=\linewidth]{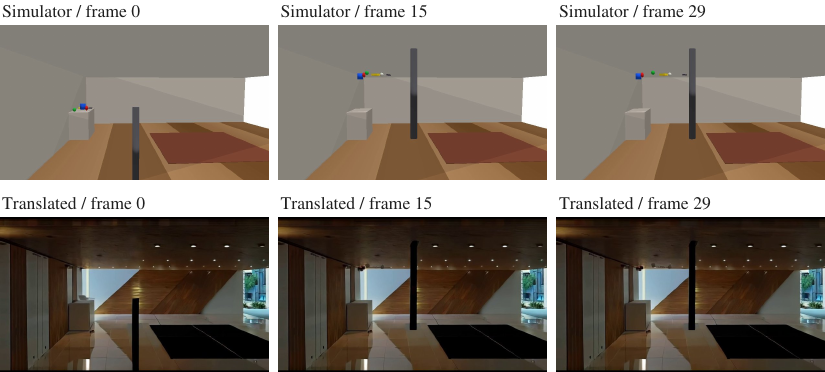}

\par\vspace{3pt}

\caption{Analytic counterfactual comparison. The displayed frames show the recorded scene; the reference answer comes from the analytic rule.}

\label{fig:example_ac}

\end{figure}

\clearpage

\begin{figure}[H]

\centering

\includegraphics[width=\linewidth]{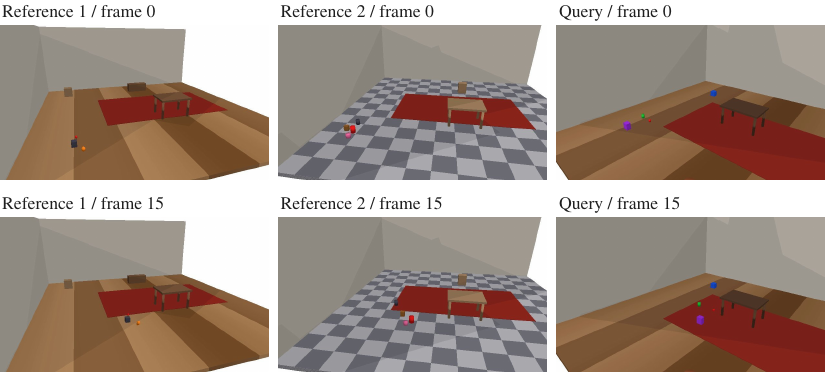}

\par\vspace{3pt}

\begin{minipage}{\linewidth}
\fontsize{9}{10.6}\selectfont\raggedright
\textbf{Question}\quad The first two videos come from the same world, whose physics differs from normal Earth physics in a specific, consistent way. Does the third video come from a world with the same physics - the same law(s) altered in the same direction and by the same amount?\par
\vspace{2pt}
\textbf{Options}\par
\begin{tabular}{@{}p{0.48\linewidth}@{\hspace{0.04\linewidth}}p{0.48\linewidth}@{}}
(1) No & (2) Yes\\
\end{tabular}\par
\vspace{2pt}
\textbf{Reference answer}\quad 1. No
\end{minipage}

\caption{Law induction with Different amount. Simulator frames show a query with a different amount category from the two reference videos.}

\label{fig:example_in_amount}

\end{figure}

\begin{figure}[H]

\centering

\includegraphics[width=\linewidth]{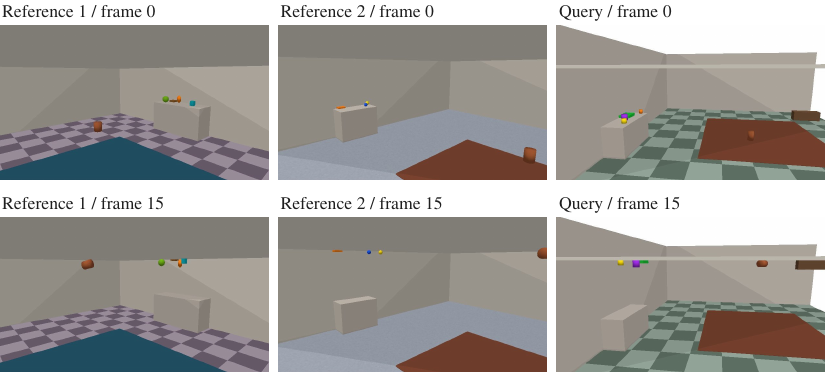}

\par\vspace{3pt}

\begin{minipage}{\linewidth}
\fontsize{9}{10.6}\selectfont\raggedright
\textbf{Question}\quad The first two videos come from the same world, whose physics differs from normal Earth physics in a specific, consistent way. Does the third video come from a world with the same physics - the same law(s) altered in the same direction and by the same amount?\par
\vspace{2pt}
\textbf{Options}\par
\begin{tabular}{@{}p{0.48\linewidth}@{\hspace{0.04\linewidth}}p{0.48\linewidth}@{}}
(1) Yes & (2) No\\
\end{tabular}\par
\vspace{2pt}
\textbf{Reference answer}\quad 1. Yes
\end{minipage}

\caption{Law induction with Match. The Simulator query video shares the reference videos' generator category.}

\label{fig:example_in_match}

\end{figure}

\end{document}